\documentclass[conference]{IEEEtran}
\IEEEoverridecommandlockouts
\usepackage{cite}
\usepackage{amsmath,amssymb,amsfonts}
\usepackage{graphicx}
\usepackage{textcomp}
\usepackage{xcolor}
\usepackage{enumitem}
\usepackage{booktabs} 
\usepackage[font=small, skip=0pt, labelsep=period]{caption}

\usepackage[linesnumbered,ruled,vlined,lined]{algorithm2e}
\usepackage{algpseudocode} 
\usepackage{subcaption}
\usepackage{arydshln}
\usepackage{array}
\def\BibTeX{{\rm B\kern-.05em{\sc i\kern-.025em b}\kern-.08em
    T\kern-.1667em\lower.7ex\hbox{E}\kern-.125emX}}
\begin{document}

\title{Split Federated Learning Over Heterogeneous Edge Devices: Algorithm and Optimization
\\
}

\author{
\IEEEauthorblockN{Yunrui Sun\textsuperscript{1}, Gang Hu\textsuperscript{1}, Yinglei Teng\textsuperscript{1}, \textit{Senior Member, IEEE}, Dunbo Cai\textsuperscript{2}}
\IEEEauthorblockA{\textsuperscript{1}Beijing Key Laboratory of Space-ground Interconnection and Convergence\\
Beijing University of Posts and Telecommunications (BUPT), Xitucheng Road No.10, Beijing, China, 100876.\\
\textsuperscript{2}China Mobile (Suzhou) Software Technology Company Limited, Suzhou,  China, 215163.\\
Email: 2022140628@bupt.edu.cn, hugang@bupt.edu.cn, lilytengtt@bupt.edu.cn, caidunbo@cmss.chinamobile.com}
}

\maketitle

\begin{abstract}
Split Learning (SL) is a promising collaborative machine learning approach, enabling resource-constrained devices to train models without sharing raw data, while reducing computational load and preserving privacy simultaneously.  However, current SL algorithms face limitations in training efficiency and suffer from prolonged latency, particularly in sequential settings, where the slowest device can bottleneck the entire process due to heterogeneous resources and frequent data exchanges between clients and servers. To address these challenges, we propose the Heterogeneous Split Federated Learning (HSFL) framework,  which allows resource-constrained clients to train their personalized client-side models in parallel, utilizing different cut layers. Aiming to mitigate the impact of heterogeneous environments and accelerate the training process, we formulate a latency minimization problem that optimizes computational and transmission resources jointly. Additionally, we design a resource allocation algorithm that combines the Sample Average Approximation (SAA), Genetic Algorithm (GA), Lagrangian relaxation and Branch and Bound (B\&B) methods to efficiently solve this problem. Simulation results demonstrate that HSFL outperforms other frameworks in terms of both convergence rate and model accuracy on heterogeneous devices with non-iid data, while the optimization algorithm is better than other baseline methods in reducing latency.  
\end{abstract}


\section{Introduction}
With the rapid expansion of the Internet of Things (IoT), vast amounts of data are being generated, offering significant potential for extracting actionable value, particularly under the driven of machine learning (ML) and deep learning (DL) techniques.  However, it may encounter significant challenges when related to privacy and security issues, making it difficult to fully realize the potential for integration and utilization of IoT data. Additionally, the heterogeneous environments created by diverse devices with varying computational resources and network conditions complicate model training and updates. To address these privacy and security concerns, as well as the challenges posed by heterogeneity, devising an effective learning framework that leverages federated and heterogeneous data while harvesting the computational resources of IoT devices presents an attractive solution\cite{MLInAir_gunduz_2019machine}.

In this regard, various distributed collaborative learning frameworks have emerged, such as federated learning (FL) \cite{mcmahan2016federated},   split learning (SL) \cite{SL_vepakomma_2018split, SL_gupta_2018distributed}. While FL has been well-studied, it still faces challenges in supporting large foundational models. In contrast, SL presents as an attractive solution by splitting the computation and preserving privacy simultaneously. SL achieves accuracy comparable to centralized learning (CL) \cite{CL}, but the sequential training process can lead to prolonged training latency and under utilization of device resources. 
Consequently, there is ongoing discussion about enabling parallel SL by introducing different aggregation process or methods. However, several challenges remain hard to tackle. \emph{First}, considering the heterogeneity in clients' storage and computational capacities, selecting the appropriate cut layer for each client to accommodate their resource constraints is crucial to prevent stragglers, which could slow down overall training latency. \emph{Second}, parallel SL requires more frequent data exchanges between clients and the server, thus, communication latency can become a bottleneck for system performance, particularly as the number of clients increases or when the channel conditions fluctuate.
\begin{figure*}[t]
\centering
\begin{subfigure}[b]{0.48\linewidth}
    \centering 
    \includegraphics[width=\linewidth]{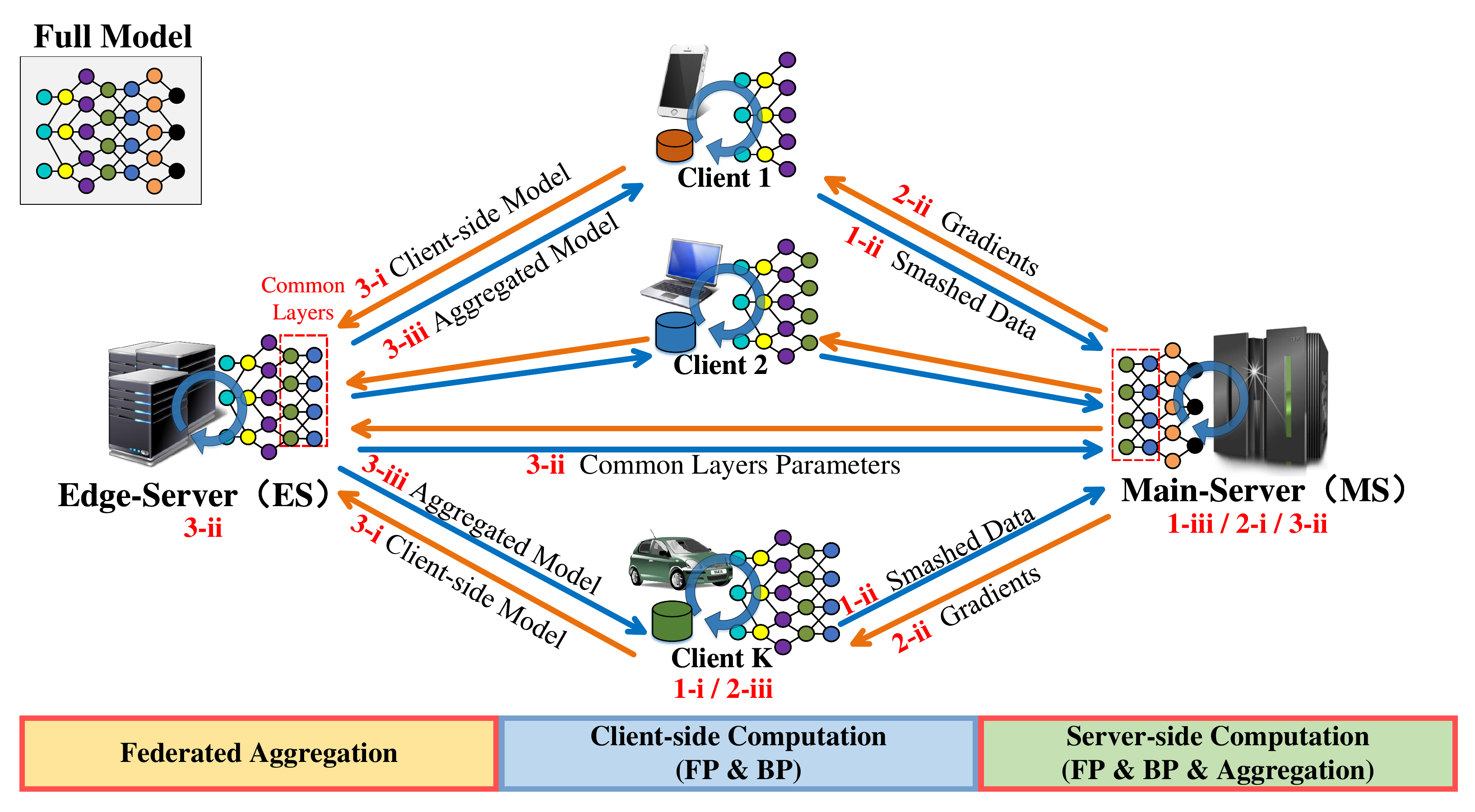}
    \caption{Framework of HSFL.}
    \label{fig:HSFL_framework}
\end{subfigure}
\hfill
\begin{subfigure}[b]{0.48\linewidth}
    \centering    
    \includegraphics[width=\linewidth]{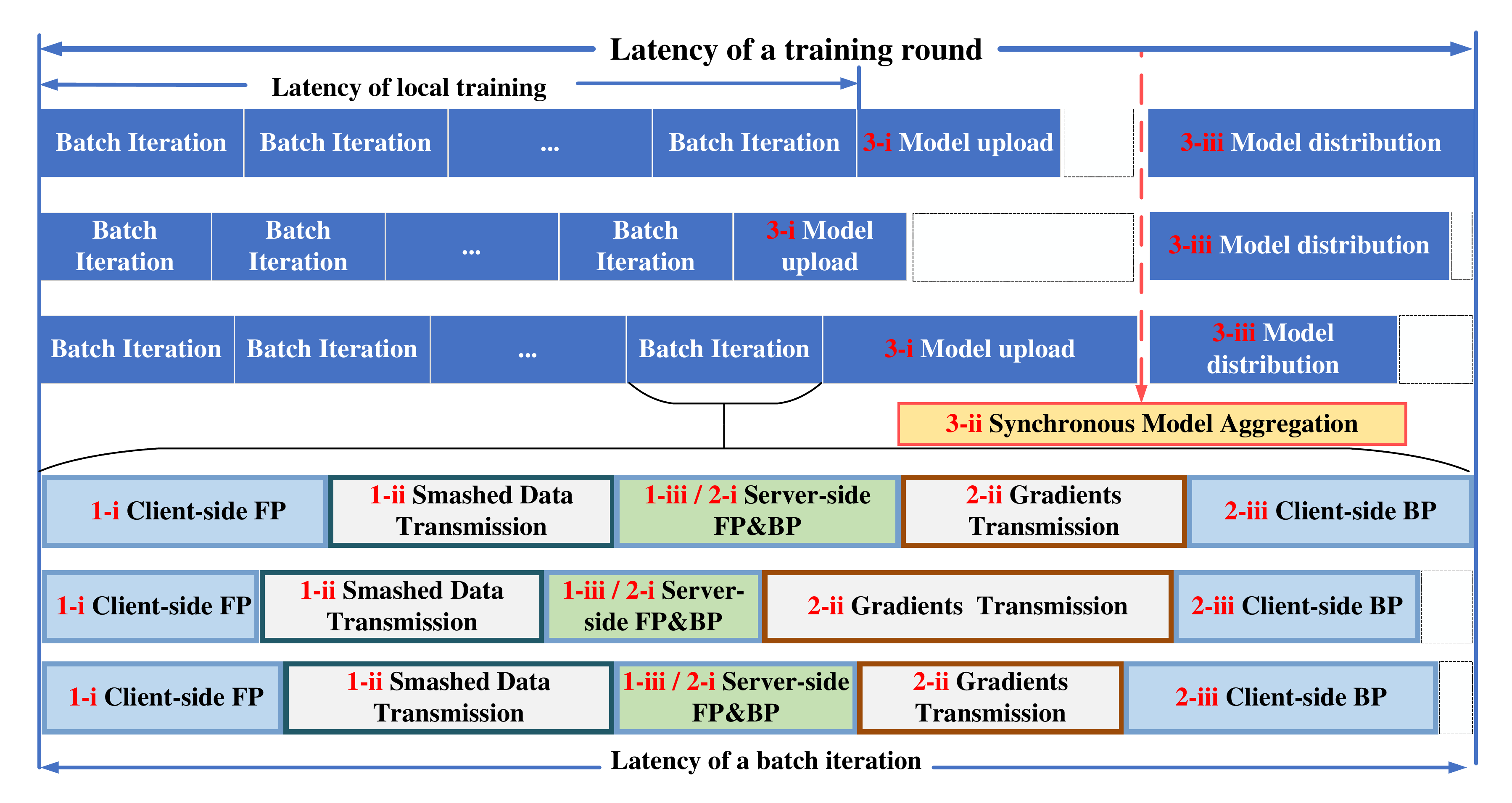}
    \caption{Single-round training latency in HSFL.}
    \label{fig:HSFL_latency}
\end{subfigure}
\caption{Framework (a) and single-round training latency (b) of HSFL. The blank in (b) means idle waiting for multi-clients.}
\vspace{-15pt} 
\label{fig1}
\end{figure*}

In existing attempts at parallel SL, the focus is primarily on achieving consistent updates for clients sharing the same cut layer, thereby maintaining global model coherence and improving training efficiency.  Jeon \emph{et al.} \cite{PSL_jeon_2020} initialize all clients with the same parameters and collects smashed data from each client to train the server-side model, returning the averaged gradients for updates. To further enhance privacy protection, Turina \emph{et al.} \cite{FSL_turina_2021} allocate an edge server for each client to store the server-side model and introduces FL to aggregate model parameters. Thapa \emph{et al.} \cite{SplitFed_thapa_2022} propose a framework, Spiltfed, which incorporates federated aggregation on both the client and server sides, blending the strengths of SL and FL. Additionally, Oh \emph{et al.} \cite{oh2022locfedmix} introduce a local-loss-based training method and combine it with above approaches, further reduces training latency. Wu \emph{et al.} \cite{CPSL} take into account the heterogeneity of client devices by clustering clients and allowing different clusters to adopt different cut layers, with parallel training within clusters
and sequential training among them. However, all of the above assume the same cut layer for splitting, neglecting the differences in computational resources and heterogeneous data among edge devices.

\addtolength{\topmargin}{0.01in}
In this paper, we propose a novel parallel SL framework, \underline{H}eterogeneous \underline{S}plit \underline{F}ederated \underline{L}earning (HSFL), that adaptively splits models on resource-constrained devices and conducts collaborative training with servers.
To the best of our knowledge, we are the first to achieve synchronous aggregation of heterogeneous models across different cut layers through the exchange of specific layers information between the servers, enabling effective training for more resource-constrained devices with non-iid data. Additionally, we formulate an optimization problem aimed at minimizing total training latency by jointly considering limited hardware capabilities and network resources. Considering the differentiated dynamics of edge status, the resulting stochastic non-convex problem is decomposed into long- and short-timescale subproblems. A Sample Average Approximation (SAA) \cite{kleywegt2002sample} -based Genetic Algorithm (GA) \cite{mirjalili2019genetic} is devised to select optimal cut layers for clients, while the Branch and Bound (B\&B) algorithm is employed to effectively allocate computational and transmission resources. Finally, simulation results demonstrate that HSFL outperforms other frameworks, with the proposed optimizations reducing training latency compared to baseline methods.

\section{HSFL Design}
As demonstrated in Fig.~\ref{fig1}(\subref{fig:HSFL_framework}), the HSFL framework involves a set of clients $\mathcal{K}=\left\{ 1,2,..., K \right\}$, a Main Server (MS) responsible for handling the computation of all the server-side models, and an Edge Server (ES) that  aggregates and distributes the client-side models. The clients possess private data and let ${{\mathsf{\mathcal{D}}}_{k}}=\left\{ {\mathbf{x}_{k,i}},{{y}_{k,i}} \right\}_{i=1}^{D_k}$ denote client $k$ local dataset, where $\mathbf{x}_{k,i}$ denotes the $i$-th sample and $y_{k,i}$ denotes the corresponding label, while $D_{k}$ represents the dataset's size. 
To enable training on resource-constrained clients, each client $k\in\mathcal{K}$ has an appropriate cut layer $l_{k}$ to fully utilize its computational and transmission resources while preventing it from falling behind during training. Then the clients and the MS cooperatively complete the model training by splitting global model $\mathbf{W}$ into client-side model 
$\mathbf{W}_{k}^{C}$ and server-side model $\mathbf{W}_{k}^{S}$ with $l_k$, and these models are deployed on client $k$ and the MS, respectively. Once all clients complete their training, the ES and the MS will collaborate to aggregate models for the next training round. Specifically, we divide the HSFL workflow into three stages to provide a clearer understanding. Let $\cal{A}$ denote the set of all training rounds, consisting of $A$ rounds, the workflow in round $a$ can be described as follows:

\subsubsection{\textbf{Forward Propagation  Stage}
\label{stage1}}
At this stage, each client uses its local dataset in parallel to complete the forward propagation (FP) in collaboration with the MS, and the process is completed through the following three steps:
\begin{enumerate}[label=\roman*)]
\item \textit{Client-side Forward Computation}: 
Each client executes FP with $\mathbf{W}_{k}^{C}$ and calculates the output as smashed data ${\mathcal{S}}_{k}$. Here, HSFL adopts Mini-batch Gradient Descent (MBGD) with mini-batch data $\mathcal{B}_{k}=\left\{\mathbf{X}_{k},\mathbf{y}_{k}\right\}$, and ${\mathcal{S}}_{k}$ is given by 
    \begin{equation} \mathsf{\mathcal{S}}_{k}=f\left( \mathbf{X}_{k};\mathbf{W}_{k,a}^{C}\right),
    \end{equation}
where $f\left(\cdot\right)$ represents the network function. 
\item \textit{Smashed Data Transmission}: 
Once completing client-side computing, the client transmits the smashed data  $\mathsf{\mathcal{S}}_{k}$ and corresponding label $\mathbf{y}_{k}$ to the MS. 

\item \textit{Server-side Forward Computation}: 
Aggregating the smashed data and labels from clients, the MS executes the remainder of FP and outputs the predicted value 
\begin{equation}\mathbf{\hat{y}}_{k}=f\left(\mathsf{\mathcal{S}}_{k};\mathbf{W}_{k,a}^{S}\right),
    \end{equation}
    thus the averaged loss of this batch is calculated as  \begin{equation}l\left(\mathcal{B}_{k};\mathbf{W}_{k,a}\right)=\frac{1}{b}{\mathcal{H}\left(\mathbf{\hat{y}}_{k},\mathbf{y}_{k}\right)},
    \label{batchloss}
    \end{equation}
    where $b$ is the batch size and $\mathcal{H}\left(\cdot\right)$ represents the cross-entropy loss function.
\end{enumerate}

\subsubsection{\textbf{Back Propagation Stage}}
At this stage, the MS and clients perform back propagation (BP) given the computed loss and updates models. Similar to forward propagation stage, there are three steps as:
\begin{enumerate}[label=\roman*)]
\item \textit{Server-side Backward Computation}: 
Given the loss in Eq. \eqref{batchloss}, the gradients are calculated by $\nabla_{\mathbf{W}_{k,a}^S}l\left(\mathcal{B}_{k};\mathbf{W}_{k,a}^S\right)$. The MS executes BP and updates $\mathbf{W}_{k,a}^{S}$ using the learning rate $\eta$: 
\begin{equation}\mathbf{W}_{k,a}^{S}\leftarrow\mathbf{W}_{k,a}^{S}-\eta{\nabla_{\mathbf{W}_{k,a}^S}l\left(\mathcal{B}_{k};\mathbf{W}_{k,a}^S\right)}.
\label{serverBP}
\end{equation}
\item \textit{Gradients Data Transmission}: 
Then MS transmits the gradients of ${\mathcal{S}}_{k}$ to the corresponding client as soon as they are computed. 

\item \textit{Client-side Backward Computation}: 
After receiving the feedback gradients, the client updates $\mathbf{W}_{k,a}^{C}$ accordingly using 
\begin{equation}\mathbf{W}_{k,a}^{C}\leftarrow\mathbf{W}_{k,a}^{C}-\eta{\nabla_{\mathbf{W}_{k,a}^C}l(\mathcal{B}_{k};\mathbf{W}_{k,a}^C)}.
\label{clientBP}
\end{equation}
\end{enumerate}

The iteration above will repeat $N$ times per epoch to completely utilize the local dataset.  
\addtolength{\topmargin}{0.01in}
\subsubsection{\textbf{Synchronous Aggregation Stage}}
At this stage, the ES and MS collaborate to aggregate the updated models, resulting in global client-side model and server-side model, which are used for the next round of training. 
\vspace{-1mm}
\begin{enumerate}[label=\roman*)]
    \item \textit{Model upload}: 
    After finishing its last local epoch, each client transmits its updated client-side model to the ES. To enable synchronous aggregation, this step may incur idle waiting across multi-clients.
    \item \textit{Federated Aggregation}: 
    Note that among the heterogeneous models due to different cut layers, specific layers (referred to as common layers in Fig. \ref{fig1}(\subref{fig:HSFL_framework})) exist in both the ES and MS. So directly applying FedAvg \cite{mcmahan2016federated} on the servers respectively would result in inconsistent aggregation for these layers. To address this challenge, we introduce an additional communication step between the ES and MS to exchange the parameters of the common layers, ensuring that both sides maintain consistent updates when applying aggregation.
    \item \textit{Model Distribution}: 
    Once aggregation is completed, the ES distributes the aggregated client-side models back to the clients according to their respective cut layers, while the global server-side model is stored in the MS for the next training round.
\end{enumerate}
\vspace{-1.5mm}
\section{System Latency Model}
\vspace{-1.5mm}
\label{training latency analysis} 
To evaluate the training efficiency of the proposed framework, we model its total training latency as illustrated in Fig.~\ref{fig1}(\subref{fig:HSFL_latency}). For ease of comparison we also analyze the latency of each stage in the workflow separately, and the detailed description is as follows:     
\subsubsection{\textbf{Latency of Forward Propagation Stage}}
    let $b\gamma_{c}^{F}(l_{k})$ and $b\gamma _{c}^{B}(l_{k} )$ denote the FP and BP computational workload (in FLOPs) with cut layer $l_{k}$. The latency of client-side forward and backward computation is given by 
    \begin{equation}
    T_{k}^{C}=\frac{b\gamma _{c}^{F}(l_{k} )+b\gamma _{c}^{B}(l_{k})}{{{f}_{k}}{{\kappa }_{k}}},
    \end{equation}
    where ${f}_{k}$ denotes client $k$'s CPU cycles per second. ${\kappa }_{k}$ is the computing intensity which indicates the CPU cycles required for one float-point operation. 
    For the transmission latency, data is transmitted over Orthogonal Frequency Division Multiple Access (OFDMA) with Time Division Duplexing (TDD), where subchannels are dynamically allocated for either uplink or downlink transmission at different times. Specifically, let $r_{k,\text{MS}}^{i}$ denote whether the channel $i$ is allocated to client $k$ or not, and let $p_{k,\text{MS}}^{i,u}$,  $g_{i,k}$ represent client $k$ 's uplink transmit power and gain on subchannel $i$, respectively. The uplink transmission rate is given by
    \begin{equation}   
    R_{k,\text{MS}}^{U}=\sum\limits_{i=1}^{I}{r_{i,\text{MS}}^{k}}B{{\log }_{2}}\left( 1+\frac{p_{i,\text{MS}}^{u}{{g}_{i,k}}}{{{\delta }^{2}}}\right),
    \label{r_ik}
    \end{equation}
    where $I$ is the number of subchannels and $B$ is the bandwidth of each subchannel, $\delta^2$ is the PSD of noise. Letting $b{\xi}_{s}({l}_{k})$  denote the data size (in bits) of the smashed data \footnote{The label size is negligible compared to the data and gradients because it typically occupies far less memory and has minimal impact on the overall communication cost.}, and its transmission latency is given by
    \begin{equation}
     T_{k,\text{MS}}^{U}=\frac{b{{\xi }_{s}}\left( {{l}_{k}} \right)}{R_{k,\text{MS}}^{U}}.
    \end{equation}
    \subsubsection{\textbf{Latency of Back Propagation Stage}\label{stage2}}
    Similarly, let $f_{s}^{k}$ represent the computing frequency that the MS allocated to client $k$'s server-side model training,  $b{\mathcal{\gamma}}_{s}^{F}({l}_{k})$ and $b{\mathcal{\gamma}}_{s}^{B}({l}_{k})$ denote the computational workload of FP and BP respectively. The computing latency on the MS is given by 
    \begin{equation}    T_{k}^{S}=\frac{b{\mathcal{\gamma}}_{s}^{F}({l}_{k})+b{\mathcal{\gamma}}_{s}^{B}({l}_{k})}{{{f}_{s}^{k}}{{\kappa }_{s}}},
    \label{Tsk}
    \end{equation}
    where ${\kappa}_{s}$ denotes the MS's computing intensity. On the transmission aspect, let $b{\xi}_g({l}_{k})$ denote the gradients' data size (in bits), the transmission latency is denoted by
    \begin{equation}
     T_{k,\text{MS}}^{D}=\frac{b{{\xi }_{g}}( {{l}_{k}})}{R_{k,\text{MS}}^{D}},
    \end{equation}
    where $R_{k,\text{MS}}^{D}$ is the downlink transmission rate whose definition is similar to Eq. \eqref{r_ik}. 
    \subsubsection{\textbf{Latency of Synchronous Aggregation Stage}}
    At this stage, the computational tasks consist of only a small amount of summation and averaging, making the computing latency negligible. For more simplicity, we also assume that the transmission of common layers between the ES and MS is efficient and minimal, thus, we neglect the latency associated with these two part. As a result, the  overall latency consists of model transmission only. Let ${\xi}_m(l_k)$ denote the client-side model size (in bits); the uplink and downlink transmission latency between clients and the ES are given by 
    \begin{equation}
    T_{k,\text{ES}}^{U} = \frac{{\xi}_m(l_k)}{R_{k,\text{ES}}^{U}},    T_{k,\text{ES}}^{D} = \frac{{\xi}_m(l_k)}{R_{k,\text{ES}}^{D}},
    \end{equation}
    where ${R_{k,\text{ES}}^{U}}$ and ${R_{k,\text{ES}}^{D}}$ are defined similarly to ${R_{k,\text{MS}}^{U}}$, ${R_{k,\text{MS}}^{D}}$.

Hence, for client $k$, the latency in a training round is $T_{k} = N(T_{k,\text{MS}} + T_{k,\text{ES}}^{U}+T_{k,\text{ES}}^{D}),
\label{client k latency}$
where $T_{k,\text{MS}}=T_{k}^{C}+T_{k}^{S}+T_{k,\text{MS}}^{U}+T_{k,\text{MS}}^{D}$\label{optimal F}. During the parallel training phase, the training latency is represented by the maximum among all clients and is given by 
\begin{equation}
\small{T_{a}\hspace{-1mm}= \hspace{-1mm} \underset{k\in\mathcal{K}}\max\left\{\min\left\{\left(N(T_{k,\text{MS}})\hspace{-1mm} +\hspace{-1mm} T_{k,\text{ES}}^{U}\right), \tau\right\}\right\} \hspace{-1mm} +\hspace{-1mm} \underset{k\in\mathcal{K}}\max\left\{T_{k,\text{ES}}^{D}\right\},
\label{per-round latency}}
\end{equation}
where $\tau$ denotes the system's latency tolerance, the minimization indicates that the system will stop waiting for  stragglers exceeding this threshold, ensuring training efficiency.

\section{Problem Formulation And Solutions}

\addtolength{\topmargin}{0.01in}
\subsection{Problem Formulation}
To improve HSFL's training efficiency, we formulate a optimization problem to minimize the total latency, jointly optimizing following variables: 1. \emph{Cut layer selections:} Denoted by $\mathbf{l} \in \mathbf{L} = \{(l_1, l_2, \ldots, l_K)\}$; 2. \emph{MS's Computing frequency allocation:} Denoted by $\mathbf{F} = (f_s^1, f_s^2, \ldots, f_s^K)$ and the element $f_s^k$ is defined in \ref{stage2}; 3. \emph{Transmission resource allocation:} To Simplify the optimization, we combine subchannel allocation and transmit power allocation into a single variable, which is denoted by $\mathbf{P} = \left(\mathbf{P}_{\text{MS}}^{u}, \mathbf{P}_{\text{MS}}^{d}, \mathbf{P}_{\text{ES}}^{u}, \mathbf{P}_{\text{ES}}^{d}\right)$, the element $p_{k,\text{MS}}^u$, $p_{k,\text{MS}}^d$, $p_{k, \text{ES}}^u$ and $p_{k,\text{ES}}^d$ represent the transmit power allocated to client $k$ in the uplink/downlink between it and the MS/ES, respectively.
So the optimization problem is formulated as follows:
\allowdisplaybreaks % 允许公式跨页断开
\begin{align}
\mathcal{P}: & \underset{\mathbf{l},\mathbf{F},\mathbf{P}}{\mathop{\min }}\,\underset{a\in\mathcal{A}}{\sum}{T_{a}}\left(\mathbf{l},\mathbf{F},\mathbf{P} \right) \\
&\!\!\!\!\!\!\!\!\! \rm{s.\;t.} \scalebox{1}{$\;\;\;\; C_1:{{l}_{k}}\in \{0,...,l_{k}^{\text{max}} \},\forall {k}\in \mathcal{K},$} \nonumber\\
&\scalebox{1}{$\;\;\;\; C_2:{f}_{s}^{k}\in (0, f_{s}),\forall k\in \mathcal{K},$}   \nonumber\\
&\scalebox{1}{$\;\;\;\; C_3:\sum\limits_{k=1}^{K}{f_{s}^{k}\le f_s},$}  \nonumber\\
&\scalebox{1}{$\;\;\;\; C_4:p_{k,\text{MS}}^u \le p_k^{\text{max}}, p_{k,\text{ES}}^u\le p_k^{\text{max}}, \forall k \in {\cal K},$}  \nonumber\\
&\scalebox{1}{$\;\;\;\; C_5:\sum\limits_{k = 1}^K {p_{k,\text{MS}}^d}  \le p_{\text{MS}}^{\text{max} },\sum\limits_{k = 1}^K {p_{k,ES}^d}  \le p_{\text{ES}}^{\text{max} },$}  \nonumber
\end{align}
where constraint $C_1$ limits selection of cut layer for each client, and $l_k^\text{max}$ denotes the largest cut layer available to client $k$ according to its storage.
$C_2$ and $C_3$ indicate the limitation of the MS's computing frequency allocation and $f_s$ denotes the MS's total computing frequency.  
$C_4$ and $C_5$ guarantee the transmit power constraints of clients and servers.

Obviously, $\mathcal{P}$ is a $\textit{stochastic}$ $\textit{non-convex}$ optimization problem. $\textit{Stochastic}$ refers to the need for decisions to take into account the temporal dynamics of client  computing ability and channel conditions throughout the training process. 
% Since these random variables fluctuate over time, the optimization process is dynamic rather than static. 
$\textit{non-convex}$ refers to the non-convex objective and constraints, making the problem NP-hard.
\vspace{-10pt}
\subsection{Problem Solution}
\begin{figure}[t]
\centering
\includegraphics[width=0.95\linewidth]{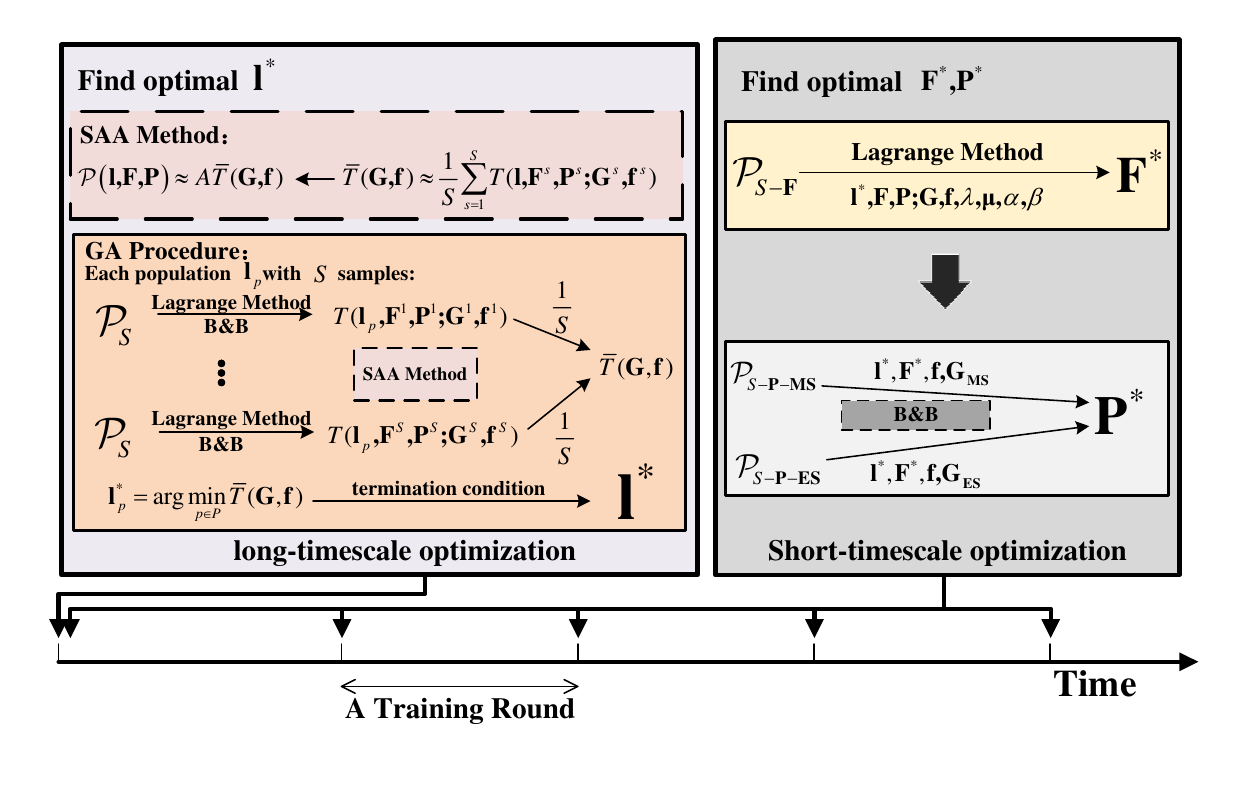}
\caption{Flow of SAA\&GA-based algorithm.}
\label{fig:algorithm relationship}
\vspace{-20pt}
\end{figure}

\vspace{-5pt}
Note that solving $\mathcal{P}$ directly is difficult, we propose an algorithm that simplifies the problem by decomposing it into manageable subproblems. Specifically, consider that the client's storage resource constraints, which indicates how large a client-side model the client can store, is stable during a training task, so the choice of cut layer $\mathbf{l}$ should be fixed for the entire training process, representing a long-timescale variable. In contrast, the allocation of computational and transmission resources, $\mathbf{F}$ and $\mathbf{P}$, should be adjusted in each training round as short-timescale variables due to the dynamic computing capability of clients and channel conditions.
%考虑这个长短期变量在哪里说，问题分析还是问题解决
In other words, we divide $\mathcal{P}$ into long-timescale subproblem $\mathcal{P}_{L}$ aiming to minimize the total training latency and short-timescale subproblem $\mathcal{P}_{S}$ focusing on optimizing the single-round training latency, which are solved separately. The optimization algorithm flow is shown in Fig. \ref{fig:algorithm relationship}.

At the beginning of training process, we solve the long-timescale subproblem $\mathcal{P}_{L}$ which is given by  
\begin{align}
\vspace{-10pt}
\mathcal{P}_{L}: & \underset{\mathbf{l}}{\mathop{\min }}\,\underset{a\in\mathcal{A}}{\sum}{T_{a}}(\mathbf{l},\mathbf{F},\mathbf{P}) \\
&\!\!\!\!\!\!\!\!\!\! \rm{s.\;t.} \scalebox{1}{$\;\;\;\; C_1-C_5,$} \nonumber
\vspace{-10pt}
\end{align}
When optimal cut layer selection $\mathbf{l}^{*}$ is obtained, the variables $\mathbf{F}$ and $\mathbf{P}$ in the $\mathcal{P}_{S}$ are still coupled, jointly influencing the single-round training latency, as shown in Eq. \eqref{per-round latency}. This presents a considerable challenge in optimizing them together. Therefore, we divide $\mathcal{P}_{S}$ into $\mathcal{P}_{S-1}$ to optimize $\mathbf{F}$ and $\mathcal{P}_{S-2}$ to optimize $\mathbf{P}$ in each training round, which are given by
\begin{align}
    \mathcal{P}_{S-\mathbf{F}}: & \underset{\mathbf{F}}{\mathop{\min }}\, T_{a}(\mathbf{l}^*,\mathbf{F},\mathbf{P}) \\
    &\!\!\!\!\!\!\!\!\!\! \rm{s.\;t.} \scalebox{1}{$\;\;\;\; C_2, C_3,$} \nonumber
\vspace{-5pt}
\end{align}
\vspace{-10pt}
\begin{align}
    \mathcal{P}_{S-\mathbf{P}}: & \underset{\mathbf{P}}{\mathop{\min }}\, T_{a}(\mathbf{l}^{*},\mathbf{F},\mathbf{P}) \\
    &\!\!\!\!\!\!\!\!\!\! \rm{s.\;t.} \scalebox{1}{$\;\;\;\; C_4, C_5.$} \nonumber
\vspace{-10pt}
\end{align}

\subsubsection{Long-timescale Variables Solution}
In $\mathcal{P}_{L}$, the challenges lies in how to reasonably integrate the dynamically changing client computing capabilities $\mathbf{f}$ and channel conditions $\mathbf{G}$ with the optimization variables to estimate the total latency. Additionally, HSFL allows clients to choose different cut layers, resulting in an exponentially large solution space, which becomes particularly significant when there are many choices. To tackle these challenges, we propose a GA algorithm based on SAA, where SAA leverages historical data samples of $\mathbf{f}$ and $\mathbf{G}$ to calculate the averaged single-round latency ${\bar{T}(\mathbf{G,f})}$ as an approximation of $T_{a}$. Then GA efficiently searches for $\mathbf{l}$ which minimize the ${\bar{T}(\mathbf{G,f})}$. 
Thus, the objective of $\mathcal{P}_{L}$ can be expressed by
\begin{equation}
\underset{a\in\mathcal{A}}{\sum}{T_{a}}\approx A{\bar{T}(\mathbf{G,f})}.
\label{total}
\end{equation}

The core steps of our SAA\&GA-based algorithm are as follows: 
$\textit{First}$, assume $\mathbf{G}$ and $\mathbf{f}$ follow Gaussian distribution with their respective means and variances and we can optimize $\mathbf{F}$ and $\mathbf{P}$ with a given  $\mathbf{l}$, $\mathbf{G}$ and $\mathbf{f}$ with solving $\mathcal{P}_{S}$, $\textit{then}$ we take $S$ historical samples of $\mathbf{G}$ and $\mathbf{f}$ to compute the single-round latencies on these samples and use their mean to approximate the expected value ${\bar{T}(\mathbf{G,f})}$, which can be described by 
\begin{equation}
{\bar{T}(\mathbf{G,f})}
    \approx \frac{1}{S} \sum_{s=1}^{S} {T}(\mathbf{l},\mathbf{F}^{s},\mathbf{P}^{s};\mathbf{G}^s,\mathbf{f}^s),
    \label{fitness function}
\end{equation}
where ${T}(\mathbf{l},\mathbf{F}^{s},\mathbf{P}^{s};\mathbf{G}^s,\mathbf{f}^s)$ denotes the optimized single-round training latency on historical sample $s$.  $\textit{Finally}$, we can find the $\mathbf{l}^*$ by GA procedure.

In other words, our proposed algorithm sets the fitness of the standard GA as the ${\bar{T}(\mathbf{G,f})}$. Since GA generally optimize towards individuals with higher fitness, we need set the fitness negative to indicate that we are aiming for minimum latency.  More specifically, we initialize a population of $P$ individuals, where the fitness for individual $l_p$ is defined by  
\begin{equation}
     \text{Fit}(\mathbf{l}_{p}) = -{\bar{T}(\mathbf{G,f})}. 
\end{equation}
Then we iterate over generations until the fitness of the best individual, which minimizes ${\bar{T}(\mathbf{G,f})}$, shows no significant improvement for $g$ consecutive generations, yielding the $\mathbf{l}^*$.  

\subsubsection{Short-Timescale Variables Solution} 
For $\mathcal{P}_{S-\mathbf{F}}$,  according to Eq. \eqref{Tsk}, \eqref{per-round latency}, given  $\mathbf{l^*}$ and corresponding $\mathbf{P}$, current $\mathbf{f}$ and $\mathbf{G}$, only $T_{k}^{S}$ in latency remains to be optimized. So the objective of $\mathcal{P}_{S-\mathbf{F}}$ can be expressed as
\begin{align}
    \mathcal{P}_{S-\mathbf{F}}: & \underset{\mathbf{F}}{\mathop{\min }}\, \underset{k\in\mathcal{K}}{\max} \left\{m_k+\frac{n_k}{f_{s}^{k}}\right\}\\
    &\!\!\!\!\!\!\!\!\!\! \rm{s.\;t.} \scalebox{1}{$\;\;\;\; C_2, C_3,$} \nonumber
    \label{non-linear}
\vspace{-5pt}
\end{align}
where $m_k=N(T_{k}^{C}+T_{k,\text{MS}}^{U}+T_{k,\text{MS}}^{D})+T_{k,\text{ES}}^{U}+T_{k,\text{ES}}^{D}$ and $n_k = \left({b{\mathcal{\gamma}}_{s}^{F}({l}_{k})+b{\mathcal{\gamma}}_{s}^{B}({l}_{k})}\right)/{{\kappa }_{s}}$ represent the values that has no relationship with $\mathbf{F}$ and can be directly calculated.

For this min-max problem with inequality constraints, direct solution is difficult. We can first introduce an auxiliary variable  $T_{m}$ that satisfies $C_6: T_{m} \geq \{m_k+\frac{n_k}{f_{s}^{k}}\}, \forall{k} \in \mathcal{K}$ to linearize the objective, and then use Lagrangian relaxation to optimize it.  Thus, $\mathcal{P}_{S-\mathbf{F}}$ can be reformulated as 
\begin{align}
\mathcal{P}_{S-\mathbf{F}}: & \underset{\mathbf{F}}{\mathop{\min }}\, T_{m} \\
&\!\!\!\!\!\!\!\!\!\! \rm{s.\;t.} \scalebox{1}{$\;\;\;\; C_2, C_3, C_6.$} \nonumber
\end{align}
Then we can construct the Lagrangian function:
\begin{align}
\vspace{-10pt}
 &\mathcal{L}\left(f_{s}^{1},...,f_{s}^{K},\lambda,\mu_{1},...,\mu_{K}\right) \nonumber=T_{m}+\lambda\left(\sum_{k=1}^{K}f_{s}^{k}-f_{s}\right)\\&+\sum_{k=1}^{K}\mu_{k}\left(m_k+\frac{n_k}{f_{s}^{k}}-T_{m}\right).
    \vspace{-10pt}
\end{align}
Given the initialized values of $\mathbf{F}$ and $\lambda$, $\mu_{k}$, the initialized objective can be calculated as $T_{m}^0$ and then we optimize it by solving the partial differential equations of $\cal{L}$: 
First, compute the $f_s^k$ for each client with $f_s^k \gets \sqrt{{\mu_k n_k}/{\lambda}}$. 
Then we can update the objective and parameters: $T_{m}^{t+1} \gets \max\limits_{k} \left\{ m_k + \frac{n_k}{f_s^k} \right\}$, $\lambda \gets \lambda - \alpha \left( \sum_{k=1}^{K} f_s^k - f_s \right)$, $\mu_k \gets \max \left( 0, \mu_k + \beta \left( m_k + \frac{n_k}{f_{s}^{k}} - T_{m}^{t+1} \right) \right)$, 
where $\alpha$ and $\beta$ are the learning rates that control the updating step sizes. The process terminates when $\left|T_{m}^{t+1} - T_{m}^{t}\right| \le \epsilon$ and we can obtain the optimal $\mathbf{F}^{*}$, where $\epsilon$ is a predefined threshold.

As shown in the aforementioned framework design, in HSFL, clients do not communicate with both the MS and ES simultaneously. Therefore, we can optimize the communication resources from the clients to the MS and ES $\left(\text{i.e. } \mathbf{P}_\text{MS} = \left(\mathbf{P}_{\text{MS}}^{u}, \mathbf{P}_{\text{MS}}^{d}\right), \mathbf{P}_\text{ES} = \left(\mathbf{P}_{\text{ES}}^{u}, \mathbf{P}_{\text{ES}}^{d}\right)\right)$  separately using the same approach. Specifically, $\mathcal{P}_{S-\mathbf{P}}$ can be further divided into 
\begin{align}
    \mathcal{P}_{S-\mathbf{P_\text{MS}}}: & \underset{\mathbf{P}}{\mathop{\min}}\, \underset{k\in\mathcal{K}}{\mathop{\max}} \left\{N(T_{k,\text{MS}})+T_{k,\text{ES}}^U\right\} \\
    &\!\!\!\!\!\!\!\!\!\! \rm{s.\;t.} \scalebox{1}{$\;\;\;\; C_4, C_5,$} \nonumber
\end{align}
\begin{align}
    \mathcal{P}_{S-\mathbf{P_\text{ES}}}: & \underset{\mathbf{P}}{\mathop{\min}}\, \underset{k\in\mathcal{K}}{\mathop{\max}} \left\{T_{k,\text{ES}}^D\right\}  \\
    &\!\!\!\!\!\!\!\!\!\! \rm{s.\;t.} \scalebox{1}{$\;\;\;\; C_4, C_5.$} \nonumber
\end{align}
For this constrained multivariable problem, the B\&B algorithm is commonly used, which decomposes the problem and calculates lower bounds for subproblems to systematically eliminate infeasible solutions and incrementally approach the optimal solution. Therefore, 
given the $\mathbf{l}^{*}$, $\mathbf{F}^{*}$, we can use the solver \text{pyscipopt} with B\&B to obtain $\mathbf{P}_{\text{MS}}^{*}$ and  $\mathbf{P}_{\text{ES}}^{*}$.

\section{Simulation And Performance Evaluation}
\addtolength{\topmargin}{0.01in}
\subsection{Simulation Setup}
We deploy the ResNet-18 \cite{he2016deep} network on HAM10000\cite{tschandl2018ham10000}, MNIST\cite{lecun1998gradient}, and CIFAR-10\cite{krizhevsky2009learning} datasets. The mini-batch size is set as $256$ and the learning rate is set as $0.001$. The system operates with a total bandwidth of $10\text{MHz}$ which are evenly distributed across $10$ subchannels. The channel gain follows the standard norm distribution with $10^{-3}\text{W}$ noise power. The transmit power of each client ranges from $[1,10]\text{W}$, and the transmit powers of MS and ES are both 100W. The computing capability of each client is uniformly distributed within $[0, 10]\times 10^{10}  \text{cycles/s}$, and the computing capability of the MS is $100\times 10^{10}  \text{cycles/s}$.

\subsection{Performance Evaluation of HSFL}
\begin{figure}
\centering
\includegraphics[width=0.62\linewidth]{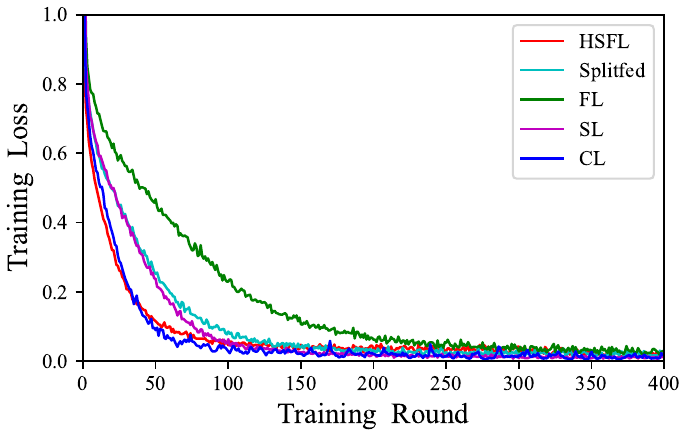}
    \caption{Convergence speed comparison on HAM10000.}
    \label{fig:ham10000_loss}
    \vspace{-15pt}
\end{figure}

To demonstrate the performance improvement, the proposed framework is compared with baselines SplitFed \cite{SplitFed_thapa_2022}, FL \cite{mcmahan2016federated}, SL \cite{SL_gupta_2018distributed}, and CL \cite{CL}.  \ref{table:acc} shows the model accuracy of those algorithms across different datasets. It is obvious that HSFL has the better performance than vanilla FL and parallel SL but slightly lower than CL. This is because our framework employs adaptive cut layers for resource-constrained device to conduct more local training. Additionally, we also compare the training convergences of these algorithms in Fig. \ref{fig:ham10000_loss}. We can see that the convergence of HSFL is close to CL but faster than other baselines. This can also be explained by the fact that HSFL achieves more local update in the same training time.
\begin{table}[h]
\centering
\vspace{-5pt}
\caption{Accuracy on different frameworks}
\label{table:acc}{
\renewcommand \arraystretch{1.3}
\begin{tabular}{lccc|cc}
    \toprule
    \textbf{Dataset} & \textbf{HSFL} & \textbf{Splitfed}& \textbf{FL} & \textbf{SL} & \textbf{CL}\\
     \textbf{HAM10000}   & \textbf{75.31} & 75.09 & 74.54 & 76.29 & 77.11 \\
     \textbf{CIFAR10} & \textbf{76.12} & 74.72 & 74.04 & 77.24 & 82.61\\
     \textbf{MNIST} & \textbf{99.05} & 98.95 & 98.27 & 98.75 & 98.93\\
\end{tabular}
}
\vspace*{-10pt}
\end{table}

\begin{figure*}[t]
\centering
\begin{subfigure}{0.30\linewidth}  % 将子图宽度设置为0.32\linewidth
\centering
\includegraphics[width=\linewidth]{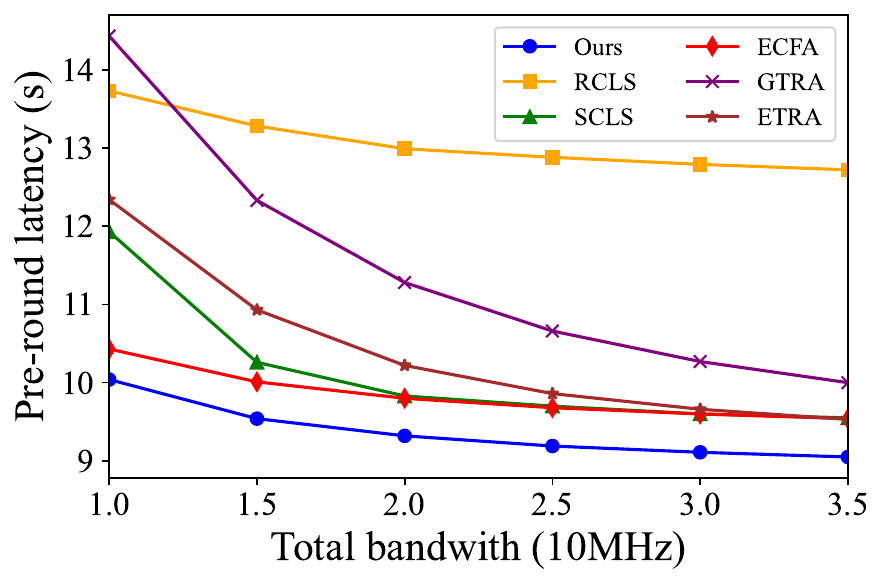}
\caption{Latency with different bandwidth.}
\label{fig:B}
\end{subfigure}
\quad  % 子图之间的间隔
\begin{subfigure}{0.30\linewidth}  % 保持每个子图宽度一致
\centering
\includegraphics[width=\linewidth]{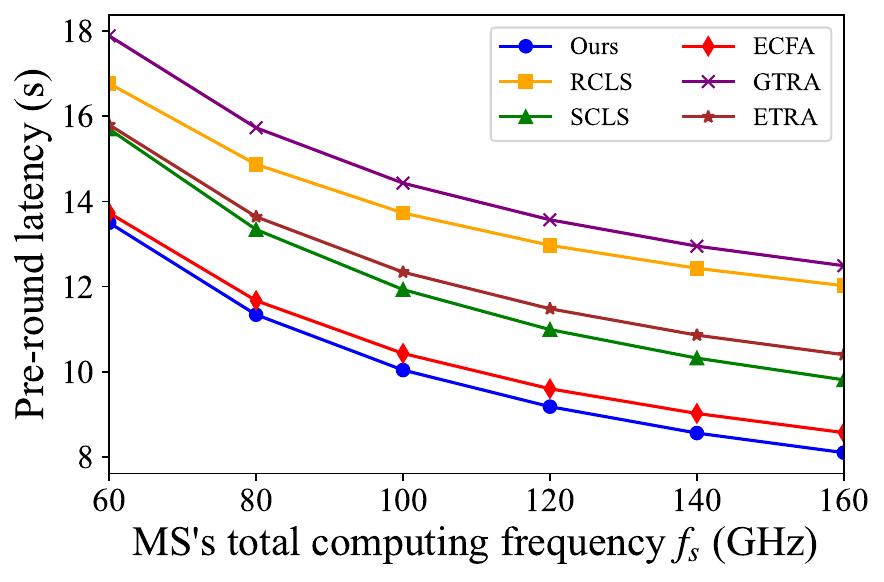}
\caption{Latency with different computing frequency.}
\label{fig:MSF}
\end{subfigure}
\quad  % 子图之间的间隔
\begin{subfigure}{0.30\linewidth}  % 保持每个子图宽度一致
\centering
\includegraphics[width=\linewidth]{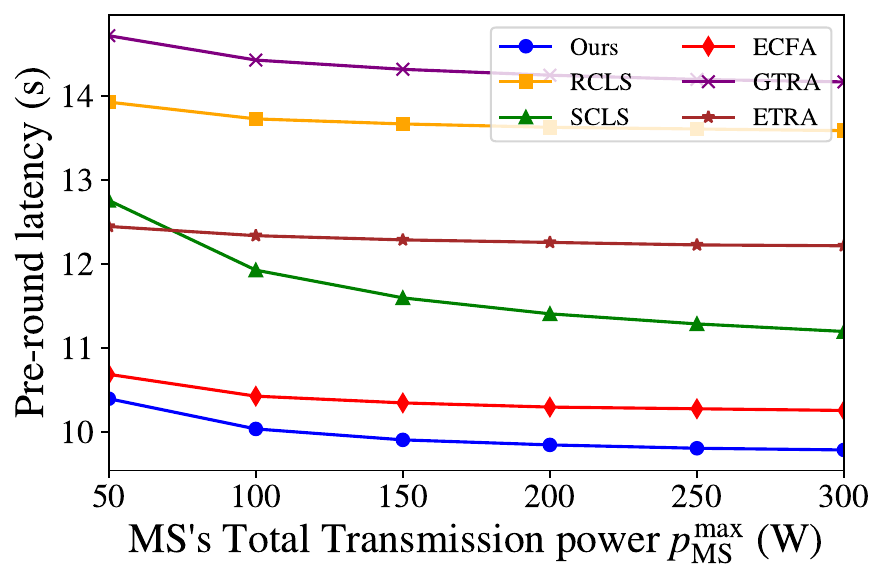}
\caption{Latency with different transmit power.}
\label{fig:MPSD}
\end{subfigure}
\caption{Single-round latency on different baselines.}
\label{different_baselines}
\vspace{-20pt} 
\end{figure*}

\addtolength{\topmargin}{0.01in}
\subsection{Performance Evaluation of the Proposed Resource Optimization Algorithm}

We compare the single-round latency among the proposed resource optimization algorithm and baselines: \textit{Rand Cut Layer Selections (RCLS)}: each client randomly selects its cut layer on feasible set. \textit{Shallowest Cut Layer Selections (SCLS)}\cite{SplitFed_thapa_2022}: system adopts the same cut layer which is minimum of the maximum cut layers for all users.  \textit{Even Computing Frequency Allocation (ECFA)}: the computing frequency of MS is allocated to clients evenly. \textit{Greedy-based  Transmission Resource Allocation (GTRA)}: the system allocate the subchannel resource with greedy method as in \cite{CPSL}. \textit{Even Transmission Resource Allocation (ETRA)}: the transmit power of MS is allocated evenly. 

As shown in Fig. \ref{different_baselines}, our proposed optimized algorithm achieves the lowest latency among all baselines in resource-constrained environment. It demonstrates the effectiveness of SAA-based GA and B\&B algorithms in resource allocation. From Fig. \ref{different_baselines}, we can see that the latency is sensitive to the change of the MS’s computing frequency. This is because the server-side computing latency significantly contributes to the total latency, especially when its computing power is not sufficiently large. Furthermore, the transmission power variation of the server has slightly impact on the overall latency, it is reasonable because the server's transmission latency constitutes only a small portion of the total latency. 

\begin{figure}
\centering
\includegraphics[width=0.6\linewidth]{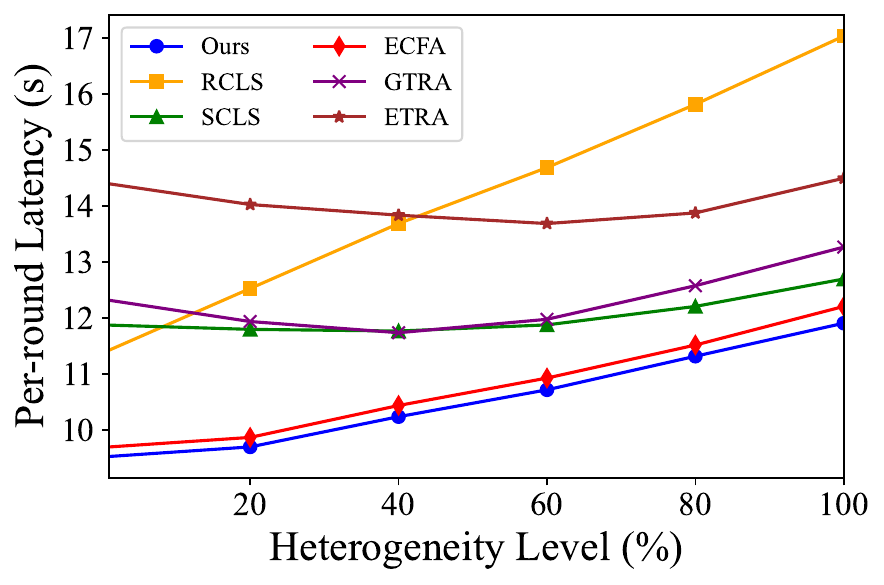}
    \caption{Single-round latency with different device heterogeneity.}
    \label{fig:comparison}
\vspace{-15pt}
\end{figure}

We simulate the impact of device heterogeneity, measured as the proportional difference in computing frequency and transmit power between devices. 
Fig. \ref{fig:comparison} shows the how the latency changes as the heterogeneity level varies. In this figure, the latency increases as the heterogeneity level increase, this is because the latency of the system is determined by the device with the most time cost. Moreover, our method outperforms other optimization algorithms, especially with high heterogeneity. This is because our method assigns each device partial model which matches its own computing capability.

\section{Conclusion}
This paper proposes a parallel SL framework, HSFL, which allows different cut layers across all clients through additional transmission between servers. We analyse the workflow and latency of HSFL and then formulate an optimization problem to minimize training latency, improving its efficiency in wireless networks. Simulation results show that HSFL outperforms existing parallel frameworks on accuracy, with a convergence speed close to CL. Additionally, our proposed resource optimization algorithm further effectively reduces training latency.

\bibliographystyle{ieeetr}
\bibliography{HSFL}
\end{document}